%% file: neurips_2020.tex
\documentclass{article}

\PassOptionsToPackage{numbers, compress}{natbib}

    \usepackage[preprint]{neurips_2020}

\usepackage[utf8]{inputenc} % allow utf-8 input
\usepackage[T1]{fontenc}    % use 8-bit T1 fonts
\usepackage{hyperref}       % hyperlinks
\hypersetup{
    colorlinks=true,
    linkcolor=blue,
    filecolor=magenta,      
    urlcolor=cyan,
}
\usepackage{url}            % simple URL typesetting
\usepackage{booktabs}       % professional-quality tables
\usepackage{amsfonts}       % blackboard math symbols
\usepackage{nicefrac}       % compact symbols for 1/2, etc.
\usepackage{microtype}      % microtypography
\usepackage[export]{adjustbox}

\usepackage{caption}
\usepackage{subcaption}
\usepackage{pgfplots}
\pgfplotsset{compat=1.16}
\usepackage{comment}
\usepackage{cancel}

\usepackage{array, makecell} 
\usepackage{xcolor}
\input{math_commands.tex}

\usepackage{authblk}
\makeatletter
\renewcommand\AB@affilsepx{, \protect\Affilfont}
\makeatother

\title{Counting Cows: Tracking Illegal Cattle Ranching From High-Resolution Satellite Imagery}

\author[1,2,5]{Issam Laradji}
\author[2]{Pau Rodriguez}
\author[3]{Freddie Kalaitzis}
\author[2]{David Vazquez}
\author[2]{Ross Young}
\author[4]{Ed Davey} 
\author[2]{Alexandre Lacoste}

\affil[1]{issam.laradji@gmail.com}
\affil[2]{Element AI}
\affil[3]{Oxford / FDL}
\affil[4]{Global Witness}
\affil[5]{McGill University}

\begin{document}

\maketitle

\begin{abstract}
Cattle farming is responsible for 8.8\% of greenhouse gas emissions worldwide. In addition to the methane emitted due to their digestive process, the growing need for grazing areas is an important driver of deforestation. While some regulations are in place for preserving the Amazon against deforestation, these are being flouted in various ways, hence the need to scale and automate the monitoring of cattle ranching activities. Through a partnership with \textit{Global Witness}, we explore the feasibility of tracking and counting cattle at the continental scale from satellite imagery. With a license from Maxar Technologies, we obtained satellite imagery of the Amazon at 40cm resolution, and compiled a dataset of 903 images containing a total of 28498 cattle. Our experiments show promising results and highlight important directions for the next steps on both counting algorithms and the data collection process for solving such challenges. The code is available at \url{https://github.com/IssamLaradji/cownter_strike}.

\end{abstract}

%%%%%%%%% BODY TEXT
\section{Introduction}
With a population nearing 1 billion, cattle are the 2nd most populous mammal on earth, after humans. Cattle farming produces 8.8\% of greenhouse gas emissions worldwide~\citep{gerber2013tackling}
\footnote{In its 2013 report, the Food and Agriculture Organization~\citep{gerber2013tackling} estimates that the livestock industry accounts for 14.5\% of anthropogenic green house gas emissions, to which, beef and dairy production contribute 40\% and 21\% respectively.}.
Besides the methane released due to digestion process of cattle, the industry demands massive regions to be deforested to make space for the grazing areas. Deforestation releases most carbon stored in the forest, it displaces indigenous communities and species, and the resulting lack of biodiversity  threatens to disrupt the ecosystem. The lack of rainforests compounds the effects of global warming, and further accelerates the ongoing 6th mass extinction of species~\cite{brook2003catastrophic, brook2006momentum, pounds2006widespread, vitousek1994beyond}.

One of the most important areas affected is the Brazilian Amazon. It is estimated that 70\% of its cleared forests are now populated by cattle, ranking Brazil's herd as the 2nd largest in the world. Despite strict laws in Brazil against the rearing and selling of cattle in protected areas of the Amazon, these are widely flouted. A common workaround to the current legislation is through smuggling the cattle to the southern grasslands, where they are declared and slaughtered in a legal area. The cattle laundering network shields cattle rearing from the law even in protected regions of the Amazon, which has been further diminished by wildfires triggered to expand the grazing areas.

This trend calls for an automated scalable way to monitor against illegal cattle ranching, a task that we show to be feasible through high-res satellite images and recent advances in deep learning. Adult cattle span 2 meters, so to achieve a reliable detection we require satellite imagery at sub-meter resolution. This could be achieved with drone imagery, but with a limited range they would be too expensive to operate on a continental-scale. On the other hand, modern satellites can be tasked for multiple, ad-hoc revisits and their on-board cameras can now resolve objects at the sub-meter level (for example, WorldView3 and SkySat constellations). In this paper we use WorldView3 (Maxar) imagery of Amazon ranches, with panchromatic (PAN) and multi-spectral (MUL) bands at 0.31 and 1.24 m/pixel respectively. Typically, MUL bands are already \textit{pan-sharpened} to the panchromatic resolution. The acquisitions from WorldView3 are task-based, so coverage can be sparse temporally and spatially, but since August 2014 most areas of the rainforest were covered a few times.

\section{Related Work}
% \paragraph{What research topics does our method intersect with?}
Our work touches on several fronts - remote sensing applied to animal conservation, localization and counting methods in ML, and environmental / human rights issues driven by deforestation.

\paragraph{Remote sensing of cattle and other animals}
% \paragraph{Remote sensing of animals}
Most of the work on the tracking and counting of animals and other small objects relies mostly on UAV/drone images~\citep{rahnemoonfar2019discountnet, weinstein2018computer}.
On cattle detection, CNNs trained with UAV images can be effective in detection~\citep{rivas2018detection, barbedo2019study} and counting~\citep{weinstein2018computer, shao2020cattle, barbedo2020counting}. 
Animals of similar size, like wildebeest, can be counted from UAV images~\citep{torney2016assessing}, and even animals as small as sheep~\citep{wang2020integrating}.
Satellite images have been broadly used for animal conservation.
Namely, WorldView2 panchromatic 50cm imagery was used to count yaks~\citep{xue2017automatic, wang2020integrating}.
However, satellite cameras hit a limit when it comes to resolving animals as small as a sheep~\citep{wang2020integrating}.
We are the first to demonstrate cattle counting from space, which contributes to the much needed evidence on the animal counting capacity of cutting-edge 0.3m sensors, made available commercially just in 2014.
% \footnote{\url{https://directory.eoportal.org/web/eoportal/satellite-missions/v-w-x-y-z/worldview-3}})

\paragraph{Localization and counting methods}
\textit{Object counting} is the task of estimating the number of objects of interest in an image. This might also include localization where the goal is to identify the locations of the objects in the image. This task has important real-life applications such as ecological surveys~\cite{arteta2016counting, saleh2020realistic, laradji2020affinity} and cell counting~\cite{cohen2017count, laradji2020weaklyWS, laradji2020weaklyAL}. The datasets used for counting tasks~\citep{guerrero2015Trancos, arteta2016counting} are often labeled with point-level annotations where a single pixel is labeled for each object. There are three main approaches to object counting: regression, density-based and detection-based methods. Regression-based methods, such as Glance~\cite{chattopadhyay2016counting}, explicitly learn to count using image-level labels only. Density-based methods~\citep{lempitsky2010learning, li2018csrnet} transform the point-level annotations to a density map through a Gaussian kernel. They are then trained to predict the density maps through with a least-squares loss. Detection-based methods~\cite{laradji2018blobs, laradji2020looc, laradji2019masks} first pinpoint the objects' locations in the images and then count the number of detected instances. LCFCN~\cite{laradji2018blobs} uses a fully convolutional neural network (FCN) and a detection-based loss function that encourages the model to output a single blob per object. In this work we evaluate a density based method, CSRNet~\cite{li2018csrnet}, and a detection method, LCFCN~\cite{laradji2018blobs}, in the task of cattle counting and localization from satellite images.

%%%%%%%%%%%%%%%%%%%%%%
% DATASET
%%%%%%%%%%%%%%%%%%%%%%
\section{Dataset}

\paragraph{Image collection process} While some satellite constellations offer global coverage with periodic revisits, Maxar's coverage is tasked-based, that is, covered regions are chosen on customer demand, and revisits occur sparsely and irregularly. As such, regions with few revisits since 2008 are commonplace. The quality of these images is far from uniform. These are compounded, for instance, by cloud coverage which varies depending on the season. The image resolution can vary, either due to an off-nadir angle of acquisition (nadir = vertical), or due to sourcing from an older sensor (e.g. WorldView2). To ensure a high overall quality, we use images with 40cm or better resolution (after pansharpening), and with less than 20\% cloud coverage. To increase the chances of observing cattle, 
%with the help of Global Witness TODO: put back after review
we geo-referenced the top selling ranches in Brazil\footnote{\href{https://www.car.gov.br/publico/imoveis/index}{Source of vector data: Brazilian Rural Environmental Register}} and queried Maxar's catalog around these locations.

% \paragraph{NDVI} Maxar's sensors provide multiple spectral bands, from which we extracted the 3 conventional Red, Green, and Blue. We also extracted the near infrared to compute an NDVI layer, which correlates with the presence of vegetation. This can provide an extra hint to both labellers and algorithms since cattle score low in this vegetation index.

% \paragraph{Pansharpened} In addition to the multi-spectral bands, a higher resolution panchromatic band is available. This high resolution monochromatic image is then used to algorithmically increase the effective resolution of the other bands. This process is called pansharpenning and enables a close to 40cm resolution. 

\paragraph{Labelling} The regions collected from Maxar's catalog were sliced into patches of $500 \times 500$ pixels. Using a custom-made labelling tool, cattle are located with point-annotations. Out of the 12,252 labelled patches, only 903 contain cattle, and the rest are labelled as ``no cow''. The distribution of the counts is shown in Figure~\ref{fig:edge_cases} (right panel). Interestingly, many of the $500 \times 500$ patches contain over 100 head of cattle, and some contain over 1000 head of cattle. In total, we have labelled 28498 head of cattle.

\paragraph{Dataset License} We plan to release this dataset for reproducibility and further machine learning development. Details of the license are still under discussion with Maxar Technologies.

%%% Figure:
% \begin{figure}
%     \centering
%     \includegraphics[width=0.32\textwidth]{figures/typical_sparse.png}
%     \caption{Typical images of cattle with various degree of crowdedness (cropped at $200 \times 200$ pixels).}
%     \label{fig:typical}
% \end{figure}

\newcommand{\maxar}[1]{Image \textcopyright 20#1 Maxar Technologies}

\begin{figure}
    \centering
    \includegraphics[width=0.24\textwidth,valign=t]{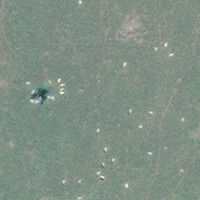}
    \includegraphics[width=0.24\textwidth,valign=t]{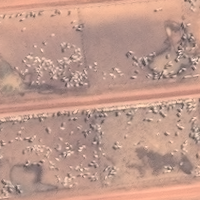}
    \includegraphics[width=0.24\textwidth,valign=t]{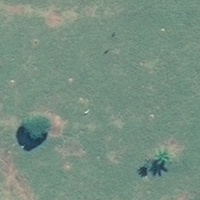}
    \includegraphics[width=0.24\textwidth,valign=t]{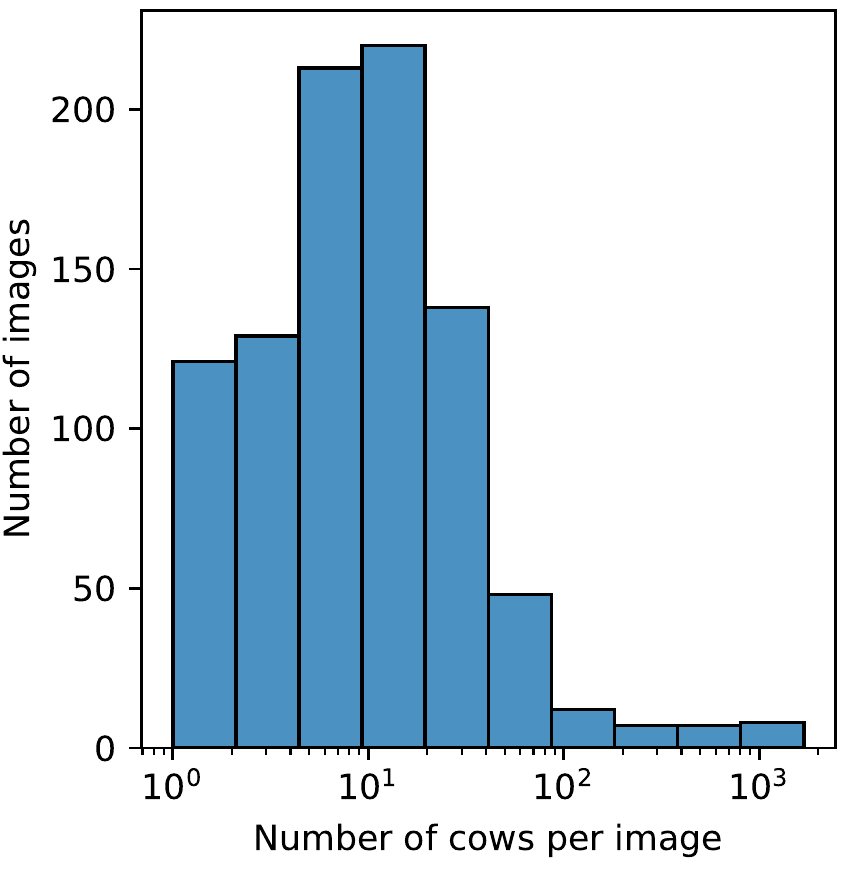}
    \caption{Peak into the dataset. \textbf{Left:} Typical image with cattle. \maxar{20} \textbf{Center left:} Crowded ranch. \maxar{20}. \textbf{Center right:} There is likely a white cow in the middle, while the rest of the white blobs appear due to lack of vegetation. \maxar{20}.
    \textbf{Right:} Histogram of the number of cattle per image.}
    \label{fig:edge_cases}
\end{figure}

\section{Cattle counting with deep learning}
We localize and count cattle in satellite images with two different approaches:

% \paragraph{Glance~\cite{chattopadhyay2016counting}} is a counting by regression method, which explicitly learns to count using image-level labels only, i.e., it only requires the overall count and not the location of the objects to count. This method is efficient when the object count is small, it is easy to train, and only requires image-level annotations. Glance is composed by a backbone network with a regression layer at the end that outputs the count. It is trained to minimize the MSE between the real and the predicted count.

\paragraph{CSRNet~\cite{li2018csrnet}} is a density-based method for counting and localization that requires point-level annotations for training. Density methods transform the point-level annotations into a density map through a Gaussian kernel. Then they are trained to predict the density maps, with a least-squares objective. Density-based methods often assume a fixed object size (defined by the Gaussian kernel), often in a controlled environment, which makes them difficult to use on objects of different shapes and sizes. These methods perform well for counting large amounts of objects in crowded situations.

\paragraph{LCFCN~\cite{laradji2018blobs}} is a detection-based method for counting and localization (LC) which can be trained with point-level annotations. It uses a fully convolutional neural network (FCN) and a detection-based loss function that encourages the model to output a single blob per object. During the training phase, the model learns to split the blobs that contain more than one point annotation and to remove the blobs that contain no point-level annotations. This method provides an accurate localization of the objects, it is robust to differences in scale, and performs well in non-crowded situations.

Our methods use a VGG16 FCN8 network~\cite{long2015fully} pretrained on ImageNet. The models are trained with a batch size of 8 for 100 epochs with ADAM~\cite{kingma2014adam} and a learning rate of $10^{-4}$. We also achieve similar results with optimizers that do not require a learning rate~\cite{vaswani2019painless, loizou2020stochastic}. We use early-stopping on the validation set and report the scores on the test set.

\section{Experiments}
In this section we assess the viability of locating and counting cattle from satellite images, as well as the weaknesses of current approaches. We compare between ImageNet pretrained-CSRNet and -LCFCN and an LCFCN with weights initialized using the Xavier method on the following two different metrics:

\newcommand{\yhat}{\hat{y}}
\paragraph{Mean Absolute Percentage Error (MAPE)} Given ground-truth counts $y$ and estimated counts $\yhat$, 
\begin{align}
    \operatorname{MAPE}= \tfrac{1}{n} \textstyle\sum_i |y_i-\yhat_i|/\operatorname{max}(y_i,1), 
\end{align}
where index $i$ runs over the evaluation set, and $\operatorname{max}$ prevents division by zero ground-truth counts. This variant of the mean absolute error (MAE) allows the comparison of crowded vs. sparse regions.

\paragraph{Grid Average Mean absolute Percentage Error (GAMPE)} measures counting and localization performance. GAMPE is a variant of GAME~\citep{guerrero2015Trancos} which uses MAPE instead of MAE. GAMPE partitions the image using a grid of non-overlapping sub-regions, and the error is computed as the sum of the mean absolute percentage errors in each of these sub-regions. This metric tends to penalize methods that rely on spurious correlations to guess the count, instead of relying on informative features.

\begin{figure}[t]
    \centering
     \includegraphics[width=0.495\textwidth]{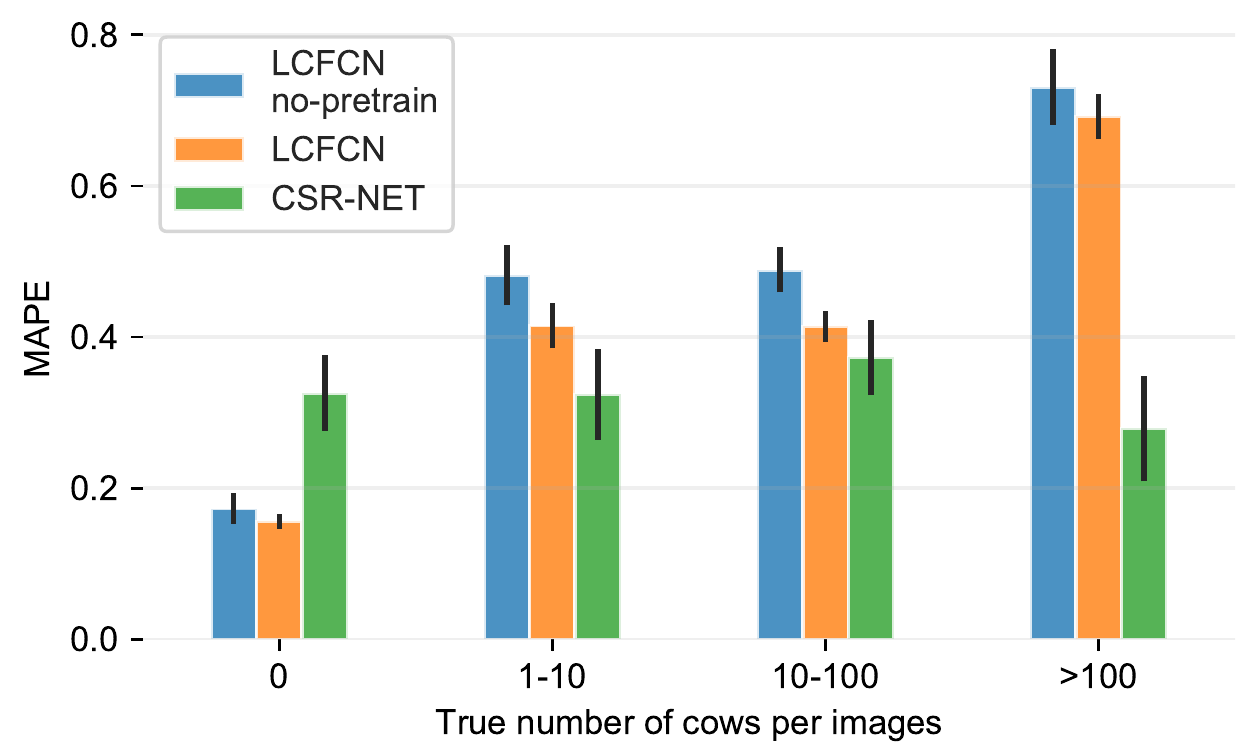}
    \includegraphics[width=0.495\textwidth]{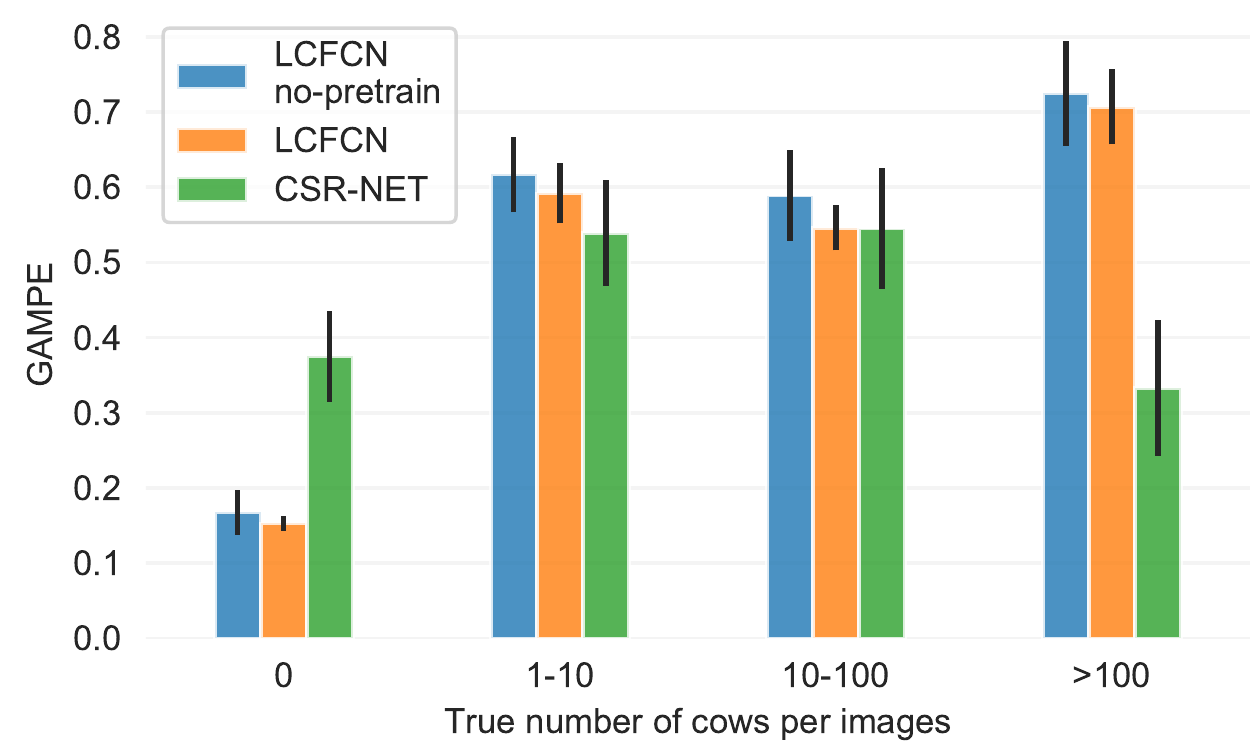}
    \caption{Counting and localization results for different cattle densities (lower is better). Error bars are $\pm 1$ std. dev. over 3 runs of the learning algorithm with different random seeds. \textbf{Left:} Mean Absolute Percentage Error (MAPE). \textbf{Right:} Grid Average Mean absolute Percentage Error (GAMPE).}
    \label{fig:results}
\end{figure}

%%% Figure
\begin{figure*}[t]
\includegraphics[width=\textwidth]{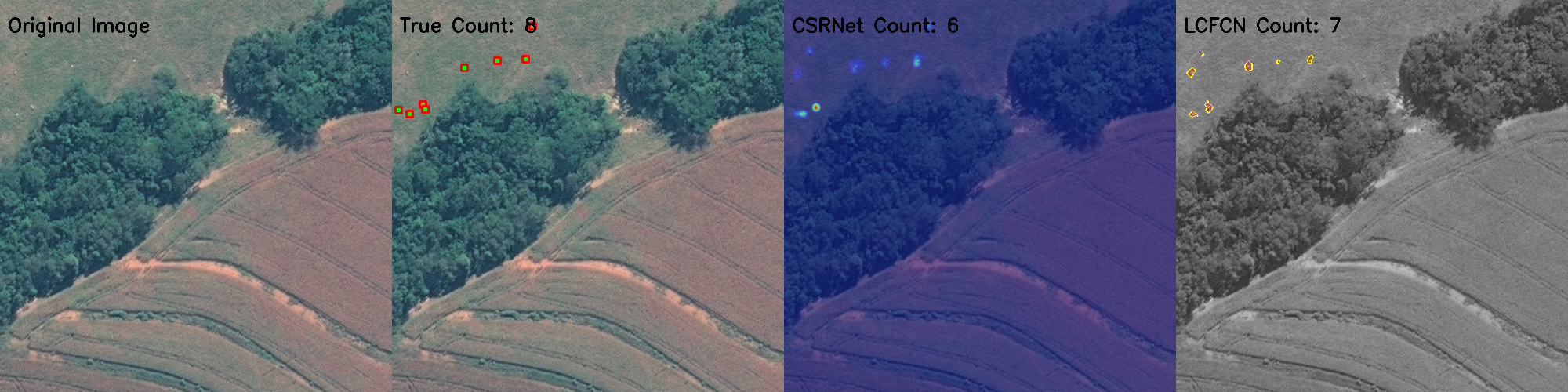}
\caption{Typical examples of results. \textbf{Left:} Original, \textbf{Center left:} Ground-truth annotation. \textbf{Center right:} CSRNet results. \textbf{Right:} LCFCN results.
\maxar{20}.}
\label{fig:qualitative}
\end{figure*}

Figure~\ref{fig:results} reports the MAPE and GAMPE on images with different cattle densities. The left-most bin (0 cattle per image) is made up of images with no cattle. In this case LCFCN outperforms CSRNet. In a binary (cattle vs no-cattle) classification benchmark, LCFCN achieves an F-score of 0.676 whereas CSRNet achieved 0.571. This result suggests that LCFCN is best suited at assessing the presence vs. absence of cattle. We also experiment with a standard ResNet-50 classifier pretrained on ImageNet, which achieves a poor F-score of 0.210 compared to LCFCN and CSRNet.

For the rest of the bins, where there are cattle present in the images (1-10 cattle per image, or more), CSRNet consistently outperforms LCFCN both on counting and localization. Note that the performance gap of CSRNet over LCFCN grows for the densest images (>100 head of cattle). We also see that LCFCN pretrained on ImageNet tends to outperform the non-pretrained LCFCN. Figure~\ref{fig:qualitative} shows a typical image from the dataset with the annotations and the results from the proposed methods.

%%%%%%%%%%%%%%%%%%%%%%
% CONCLUSIONS
%%%%%%%%%%%%%%%%%%%%%%
\section{Conclusion}
We showed that cattle detection from modern high-resolution satellite sensors is achievable, but further work is needed for a large-scale deployment. Our results suggest that a hybrid model of LCFCN and CSR-NET has merit. However, we believe that the biggest improvement would come from incorporating more information. Specifically, if we use different revisits of the same geo-location, it would be easier to distinguish cattle from static objects that resemble cattle, like bushes, rocks, and patches of sand. This would enhance the labelling quality and the counting algorithm.

%%%%%%%%%%%%%%%%%%%%%%
% BROADER IMPACT
%%%%%%%%%%%%%%%%%%%%%%
\section*{Broader Impact}
The same methodology can be applied for tracking any large animals that span more than 1 meter from the sky. This could be valuable for the census of wildlife. Ecologists could monitor animal populations more frequently, at lower costs, and act faster in the face of a species extinction.

On the flip side, high-resolution satellite imagery can be abused for surveillance\citep{santos2019satellite}. We believe that the main limiting factor to such applications is the access to high-resolution images, and not the development of counting algorithms. Also, the identification of a specific individual requires resolutions that are not yet available in commercial satellites.

%%%%%%%%%%%%%%%%%%%%%%
% ACKNOWLEDGES
%%%%%%%%%%%%%%%%%%%%%%
% \begin{ack}
%   ...
% \end{ack}

%%%%%%%%%%%%%%%%%%%%%%
% REFERENCES
%%%%%%%%%%%%%%%%%%%%%%
{\small
\bibliographystyle{abbrvnat}
\bibliography{neurips_2020}
}

\end{document}

%% file: math_commands.tex
\usepackage{amsmath,amsfonts,bm}

\def\eqref#1{equation~\ref{#1}}
\def\1{\bm{1}}

\DeclareMathAlphabet{\mathsfit}{\encodingdefault}{\sfdefault}{m}{sl}
\SetMathAlphabet{\mathsfit}{bold}{\encodingdefault}{\sfdefault}{bx}{n}